\documentclass[12pt]{article}
\usepackage{amsmath,amssymb,geometry}
\usepackage{authblk}

\usepackage[round,authoryear]{natbib} 
\title{Reverse Supervision at Scale: Exponential Search Meets the Economics of Annotation}

\author{Masoud Makrehchi\thanks{Corresponding author: \texttt{<masoud.makrehchi@ontariotechu.ca>}}}
\affil{Department of Electrical, Computer \& Software Engineering,\\
Ontario Tech University, Oshawa, Ontario, Canada}
\date{}
\begin{document}
\maketitle

\begin{abstract}
We analyze a reversed-supervision strategy that searches over labelings of a large unlabeled set \(B\) to minimize error on a small labeled set \(A\). The search space is \(2^n\), and the resulting complexity remains exponential even under large constant-factor speedups (e.g., quantum or massively parallel hardware). Consequently, arbitrarily fast, but not exponentially faster, computation does not obviate the need for informative labels or priors. In practice, the machine learning pipeline still requires an initial human contribution: specifying the objective, defining classes, and providing a seed set of representative annotations that inject inductive bias and align models with task semantics. Synthetic labels from generative AI can partially substitute provided their quality is human-grade and anchored by a human-specified objective, seed supervision, and validation. In this view, generative models function as \emph{label amplifiers}, leveraging small human-curated cores via active, semi-supervised, and self-training loops, while humans retain oversight for calibration, drift detection, and failure auditing. Thus, extreme computational speed reduces wall-clock time but not the fundamental supervision needs of learning; initial human (or human-grade) input remains necessary to ground the system in the intended task.
\end{abstract}

\section{Introduction}
Supervised learning remains the dominant paradigm in modern machine learning, often estimated to account for the vast majority of practical systems in use. In supervised settings, models learn a mapping from inputs to outputs using examples where the correct answer is known in advance \citep{chapelle2006ssl,jordan2015mlperspectives,lecun2015deep,domingos2012few}. This dominance is not accidental: many real-world tasks, spam filtering, medical image triage, speech recognition, credit scoring, are naturally framed as predicting labeled outcomes, and their success depends on how well labels capture the task’s semantics. Even when researchers propose new architectures or training tricks, these methods typically achieve their strongest results when anchored by high-quality labeled datasets that clearly express the target behavior.

In day-to-day work, the bottleneck is rarely just “which model should we pick?” Practitioners consistently spend more hours understanding data, cleaning errors, reconciling schema changes, handling missing values, balancing classes, and defining consistent label guidelines than they do on model design \citep{sambasivan2021datacascades,sculley2015techdebt}. A useful shorthand is

\begin{equation}
    \text{ML}=\text{Algorithm}+\text{Human-encoded Data}.
\end{equation}

The algorithm supplies capacity and inductive bias, but the human-encoded data-curated examples, labeling rules, and documentation—supplies meaning. Annotation is therefore the heart of data encoding: each labeled example is a compact statement of intent, telling the model what counts as correct under real operational constraints. When annotation is sloppy or inconsistent, models faithfully learn that confusion.

Because labeling is costly, teams turn to strategies that make each label count. Active learning does this by choosing which examples to annotate next, typically aiming for the most informative samples, uncertain, diverse, or high-impact points that reduce error fastest. This approach can cut labeling budgets while improving coverage of rare but important cases. Still, many projects face a “cold-start” phase with little or no data. Here, practitioners often import knowledge from related problems (transfer learning), adapt a pre-trained model to the new task with a small set of labels (fine-tuning), or train a shared representation that supports several tasks at once (multi-task learning). These methods do not eliminate the need for labeled data; instead, they reduce the amount required to reach acceptable performance and speed up convergence by injecting useful priors learned elsewhere.

Section~2 develops a cost-centric view of supervision, shifting the focus from dataset \emph{size} to dataset \emph{cost}. It presents a practical blueprint in three parts—\emph{Reduce}, \emph{Reuse}, and \emph{Recycle}—covering semi-supervised/active learning, transfer/adaptation, and weak/distant supervision with synthetic augmentation, and then synthesizes these pieces into an end-to-end pipeline with governance and validation. Section~3 formalizes the reversed-supervision setup for binary classification, defines the evaluation signal on the trusted labeled set, and frames supervision itself as the decision variable (a search over labelings of the unlabeled pool). Section~4 analyzes the induced combinatorial search space, establishing exponential upper bounds and clarifying why heuristics without structural guarantees do not change worst-case behavior. Section~5 examines the limits of ultra-high-performance hardware: it derives the $O(2^n/L)$ runtime under generic speedups, validates the asymptotics across constant/polynomial/exponential regimes, and discusses quantum-specific considerations (Grover-type bounds and QRAM/oracle costs), concluding that faster hardware reduces wall-clock time but not the exponential scaling. Section~6 concludes with implications for practice, emphasizing structure and human (or human-grade) input as the primary levers for sustainable accuracy, safety, and alignment, and outlining directions for future work.

\section{Reducing the Cost of Training Data}
\label{size}
The last decade of machine learning has celebrated scale: bigger datasets, larger models, and longer training runs. Yet scale conceals a stubborn constraint: \emph{cost}. Labeled examples incur monetary expense (annotation wages, expert time), organizational drag (guideline design, QA, governance), and opportunity costs (time-to-market, reallocation of scarce domain experts). Even unlabeled data are not free once we account for acquisition, storage, compliance, and curation. Accordingly, the central question is shifting from ``How many training examples are enough?'' to ``How much training data \emph{cost} is too much for the marginal gain we obtain?'' This reframing encourages cost-aware learning strategies that trade raw quantity for structure, reuse, and automation.

We can formalize the objective by separating accuracy from expenditure. Let $Q(\mathcal{D})$ denote a quality metric (e.g., validation accuracy, utility under task loss) achieved with dataset $\mathcal{D}$, and let

\begin{equation}
    \mathrm{Cost}(\mathcal{D}) \;=\; \mathrm{Cost}_{\text{label}}+\mathrm{Cost}_{\text{curate}}+\mathrm{Cost}_{\text{compute}}+\mathrm{Cost}_{\text{latency}}+\mathrm{Cost}_{\text{risk}},
\end{equation}

where labeling covers human annotation and adjudication; curation covers acquisition, cleaning, and compliance; compute includes training and iteration cycles; latency reflects time-to-deploy; and risk aggregates expected costs from bias, safety, and model drift. A cost-aware objective is then to maximize \emph{performance per dollar}, e.g.,
\begin{equation}
\max_{\mathcal{D},\,\theta}\;\; \frac{Q(\mathcal{D},\theta)}{\mathrm{Cost}(\mathcal{D})},
\end{equation}
or, equivalently, to meet a target $Q^\star$ while minimizing total cost. Under this lens, the lever is not only \#examples but \emph{how we obtain and use them}. We highlight three complementary strategies, \emph{Reduce, Reuse, Recycle}, that lower effective training-data cost without sacrificing alignment to the task.

\subsection{Reduce: Semi-Supervised and Cost-Sensitive Labeling}
Reduction aims to get more signal from fewer labels by exploiting structure in unlabeled data. Semi-supervised learning (SSL) couples a small labeled core with a large unlabeled pool via consistency regularization, entropy minimization, or pseudo-labeling \citep{grandvalet2005entropy,lee2013pseudolabel,tarvainen2017meanteacher}. The key intuition is that the decision boundary should be smooth under realistic perturbations: if an unlabeled example is confidently classified, we can treat its prediction as a temporary label and train against it, iteratively improving coverage at marginal labeling cost. Active learning further reduces cost by \emph{selecting} which items to annotate: uncertainty sampling, core-set selection, or disagreement-based criteria identify examples with the highest expected utility, converting each human click into maximal error reduction \citep{settles2009alSurvey,roy2001expected}. In practice, teams combine SSL with active learning and rigorous QA, yielding steep improvements along the ``quality-per-label'' curve, especially for long-tail phenomena where na\"{i}ve random labeling is wasteful.

\subsection{Reuse: Transfer Learning and Adaptation}
Reuse leverages prior work to amortize supervision. Pretraining on related tasks or broad corpora distills general features that require only light adaptation for a new objective. Fine-tuning adjusts a pretrained backbone with a small task-specific set, while domain adaptation corrects shifts between source and target distributions (feature alignment, adversarial adaptation, or data augmentation that matches target statistics) \citep{long2015dan}. Multi-task learning shares representations across sibling tasks, spreading annotation costs while improving generalization through inductive transfer. Foundation and instruction-tuned models extend this logic at scale: they encapsulate vast prior supervision and can be steered with modest task-specific signals (few-shot prompts, small LoRA or adapter updates), substantially lowering the marginal dollars per unit of additional performance \citep{houlsby2019adapters,hu2022lora}.

\subsection{Recycle: Distant/Weak Supervision and Prompt Engineering}
Recycling converts \emph{non-traditional} supervision into training signal. Distant and weak supervision use heuristics, programmatic rules, knowledge bases, or multiple noisy labeling functions to generate \emph{noisy} labels at scale; label models then estimate and correct for noise to produce denoised training sets. Prompt engineering and synthetic data generation (via large generative models) produce labeled exemplars on demand, which can seed or augment scarce classes. Critically, these approaches require governance to control compounding bias and hallucination: human-in-the-loop audits, calibration checks, and holdout evaluation on verified gold data. In a mature pipeline, generative models serve as \emph{label amplifiers}, not unconditional label sources, turning a small human-curated core into a much larger supervised corpus with traceable provenance \citep{ratner2016dataprogramming,bach2017labelmodels,mintz2009distant,wang2023selfinstruct,guo2017calibration,mitchell2019modelcards}.

\subsection{Cost-centric ML Pipeline}
A cost-centric pipeline typically interleaves these strategies: begin with a compact, high-quality seed set; \emph{reduce} through SSL and active selection; \emph{reuse} via transfer from an appropriate backbone; and \emph{recycle} weak/distant signals and synthetic examples under human oversight. Iterative evaluation on trusted validation sets enforces guardrails against drift and quality regression. The result is not necessarily fewer data points, but fewer \emph{expensive} data points, with more of the supervision burden shifted to structure, prior knowledge, and targeted human judgment, exactly where human time buys the most model improvement per dollar.

Even with ultra-high-performance hardware, including hypothetical quantum accelerators—the central question persists: can raw compute permanently retire the human cost of supervision? Our analysis argues no. Faster machines shrink wall-clock time but do not overturn the exponential search inherent in unstructured label assignment, nor do they resolve the deeper issue that labels encode \emph{intent} and \emph{semantics} that must be specified somewhere. At best, extreme compute and generative models act as powerful \emph{label amplifiers}, propagating a small, high-fidelity human (or human-grade) core across vast unlabeled corpora under active, semi-supervised, or self-training loops. But without an initial, trusted specification, and ongoing human oversight to detect drift, adjudicate edge cases, and calibrate failure modes, the system risks optimizing the wrong objective at scale. Thus, the investigation that follows evaluates whether hardware alone can erase human cost “once and for all,” or merely repositions that cost toward smaller seed sets, sharper priors, and cheaper—but not zero—supervision.

\section{Reversed Supervision}





We start from a binary classification setting with two datasets of very different character. The first, $A$, is a small labeled sample of size $m$. Each element in $A$ pairs an input with its correct class, so $A$ anchors the task semantics. The second, $B$, is a much larger pool of unlabeled inputs of size $n$, with $n \gg m$. In many practical problems this imbalance is the norm: labeled examples are scarce and expensive, while unlabeled data are abundant but mute about the target concept. If we follow the most straightforward recipe, train on $A$ and evaluate on $B$, performance is typically disappointing. With so few labels, the model tends to overfit peculiarities of $A$, failing to generalize to the broader distribution represented by $B$. Put simply, $A$ does not adequately span the decision boundary or the variation present in $B$, suggesting that more labeled data would help—but at a cost that motivates the core question: can we reduce or avoid additional human annotation?

The conventional pipeline can be written as
\begin{equation}
    A \;\longrightarrow\; \text{train} \;\longrightarrow\; \text{model} \;\longrightarrow\; \text{test on } B.
\end{equation}
To challenge its limitations, consider a reversed strategy that treats the abundant set $B$ as if it were the training resource. Because $B$ lacks labels, we first assign labels to some or all of its items, potentially at random or according to a heuristic, creating a pseudo-labeled training set. We then train a model on this surrogate data and measure its quality not on $B$ (which would reward consistency with our own guesses) but on the small trusted set $A$. Let $\mu$ denote the error computed on $A$; $\mu$ is our objective signal because $A$ carries ground truth.

Under this formulation, the learning problem becomes an explicit search over possible labelings of $B$ to minimize the evaluation error on $A$. Formally, we seek a labeling function $\ell: B \to \{0,1\}$ such that, after training on $\{(x,\ell(x)) : x\in B\}$, the resulting model achieves the smallest $\mu$ when tested on $A$. This reframing elevates labeling from a fixed input to a variable to be optimized, turning supervision itself into the decision variable. The appeal is clear—if we could find a “good” labeling of $B$, we would unlock the value of abundant unlabeled data without paying for human annotation up front. The challenge, explored in the rest of the paper, is that the space of labelings is vast and the signal $\mu$ comes from a tiny set $A$; without additional structure or priors, the search is fragile and easily biased.




\section{Complexity Analysis}
At the core, the reversed-supervision strategy turns learning into an explicit optimization over label assignments for the unlabeled pool $B$. Let $\mu$ denote an evaluation metric computed on the trusted set $A$ (e.g., error, loss, or $1-\text{accuracy}$). The objective is to find a labeling function $\ell: B \to \{0,1\}$ that minimizes $\mu$  after training on the pseudo-labeled set $\{(x,\ell(x)) : x \in B\}$. This elevates supervision itself to the decision variable: the quality of the final model is now a function of the label configuration on $B$, and the optimization landscape is induced by how those labels shape the model's fit and generalization to $A$.

Because each of the $n$ items in $B$ may independently receive one of two labels, the number of distinct labelings is
\begin{equation}
    \#\{\text{labelings of }B\} = 2^n.
\end{equation}
In the absence of additional structure or priors, there is no a priori reason to expect that a random labeling is close to optimal; the probability that an uninformed guess coincides with the best labeling is
\begin{equation}
    \mathbb{P}(\text{hit optimum by chance}) = \frac{1}{2^n}.
\end{equation}
Consequently, any exhaustive approach that guarantees the global optimum must, in principle, consider all $2^n$ labelings, retraining and reevaluating a model for each candidate configuration. This is the hallmark of combinatorial explosion: the search space doubles with every additional unlabeled example.

Let $t_c$ denote the time required to train and evaluate one model (train on $B$ with a proposed labeling and test on $A$ to compute $\mu$). An exhaustive procedure incurs a total time
\begin{equation}
    T_c(n) = 2^n \cdot t_c,
\end{equation}
which implies the asymptotic upper bound
\begin{equation}
    T_c(n) = O(2^n).
\end{equation}
In practice, heuristics (e.g., greedy relabeling, relaxation to continuous surrogates, or meta-learning priors) can prune the search, but without structural guarantees they do not change the worst-case exponential character of the problem. This analysis underscores why purely label-search-based supervision is computationally prohibitive at scale and why additional inductive biases (human annotation, constraints, or domain priors) are essential to make the problem tractable.



\section{Ultra-high Performance Computing Case}
Assume access to an ultra-high-performance computer that can complete a single train-test cycle in time $t_q$ (per candidate labeling of $B$). Define the speedup factor relative to a classical implementation as
\begin{equation}
    L \;=\; \frac{t_c}{t_q},
\end{equation}
so that $L>1$ indicates a faster machine. Because the label-search procedure must still range over the $2^n$ possible assignments in the worst case, the total runtime on this faster machine is
\begin{equation}
    T_q(n) \;=\; 2^n \cdot \frac{t_q}{1} \;=\; 2^n \cdot \frac{t_c}{L}
\;=\; \frac{2^n}{L}\, t_c
\quad\Rightarrow\quad
T_q(n) \in O\!\left(\frac{2^n}{L}\right).
\end{equation}
This expression makes explicit that the machine's raw speed enters only as a multiplicative factor $1/L$ in front of the combinatorial term $2^n$. Unless $L$ itself grows \emph{exponentially} in $n$, the exponential character of the search remains.

\paragraph{Asymptotic validation.}
Consider three representative regimes for $L$:
\begin{align*}
\text{(i) Constant speedup:} \quad & L = \Theta(1) 
\;\;\Rightarrow\;\; T_q(n) = \Theta(2^n), \\
\text{(ii) Polynomial speedup:} \quad & L = \Theta(n^\alpha),\; \alpha>0
\;\;\Rightarrow\;\; T_q(n) = \Theta\!\left(\frac{2^n}{n^\alpha}\right) = \Theta(2^n), \\
\text{(iii) Exponential speedup:} \quad & L = \Theta(2^{\beta n}),\; 0<\beta\le 1
\;\;\Rightarrow\;\; T_q(n) = \Theta\!\left(2^{(1-\beta)n}\right).
\end{align*}
Only regime (iii) changes the complexity class; even then, unless $\beta=1$ (which would unrealistically cancel the entire $2^n$ factor), the runtime remains exponential in $n$ albeit with a smaller exponent. Thus, constant or polynomial improvements in hardware speed do not convert the worst-case search into a polynomial-time procedure.

\paragraph{Quantum-specific considerations.}
In the idealized setting of \emph{unstructured} search over $N$ candidates, Grover's algorithm yields a quadratic speedup $O(\sqrt{N})$ queries instead of $O(N)$. If we identify $N=2^n$ labelings, the query complexity becomes $O(2^{n/2})$, i.e.,
\begin{equation}
    T_{\text{Grover-like}}(n) \in O\!\left(2^{n/2}\right),
\end{equation}
which is an exponential improvement in the exponent but still not polynomial in $n$. Moreover, mapping each candidate labeling to a train-test outcome requires a problem-specific quantum oracle and data access model (e.g., QRAM) that may dominate or even negate the theoretical speedup. In short, known quantum speedups for unstructured search reduce, but do not eliminate, the exponential dependence arising from the $2^n$-sized hypothesis space \citep{grover1996search,boyer1998tight,giovannetti2008qram,aaronson2015fineprint}.

\paragraph{Implication.}
The expression $T_q(n)=O\!\left(2^n/L\right)$ formally captures the best-case effect of raw computational acceleration on the label-search paradigm: faster hardware reduces wall-clock time by a factor of $L$ but leaves the exponential scaling intact unless $L$ itself grows exponentially with $n$. Consequently, even with hypothetical ultra-fast (including quantum) computation, one still requires additional \emph{structure}, human-provided labels, constraints, priors, or problem reductions, to avoid the worst-case exponential search over labelings.

\section{Conclusion}
We formalized reversed supervision as an optimization over labelings of the unlabeled pool \(B\) and showed that the induced search space scales as \(2^n\). Even with ultra-high-performance hardware, including quantum-style gains or massive parallelism, the best-case improvement is multiplicative, not a change in complexity class, unless the speedup itself grows exponentially with \(n\). Known quantum results for unstructured search (e.g., Grover) reduce the exponent but do not yield polynomial time, and practical oracle/QRAM requirements introduce additional bottlenecks. Thus, compute alone cannot retire the supervision burden ``once and for all.''

What does alter the economics is \emph{structure}: human-specified objectives, task taxonomies, label schemas, constraints, and compact gold sets that inject inductive bias. Cost-effective pipelines deploy human effort surgically: seed sets for grounding; active learning for label prioritization; semi-supervised and self-training to leverage unlabeled data; transfer learning and adapters to reuse prior supervision; and weak/distant supervision plus synthetic data to recycle non-traditional signals. Generative models act as \emph{label amplifiers} anchored to a trusted human core and validated on held-out gold data.

In short, ultra-fast hardware accelerates iteration but does not eliminate fundamental supervision needs. Accuracy, safety, and alignment remain tied to human (or human-grade) input at initialization and whenever tasks or distributions shift. Future work should develop principled supervision cost models, robust validation under shift, and hybrid pipelines that combine small amounts of high-value human input with automated amplification to keep total cost bounded without sacrificing reliability.


\end{document}